\documentclass{article}
\usepackage{spconf,amsmath,graphicx}
\usepackage{multirow}
\usepackage[table]{xcolor}
\usepackage{csquotes}

\newcommand{\st}{\textcolor{purple}}
\newcommand{\asr}{\textcolor{teal}}

\title{Token-Level Serialized Output Training for Joint Streaming ASR and ST Leveraging Textual Alignments}
\name{Sara Papi$^{\ddag}$\sthanks{Work done during an internship at Microsoft.}, Peidong Wang$^\dag$, Junkun Chen$^\dag$, Jian Xue$^\dag$, Jinyu Li$^\dag$, Yashesh Gaur$^\dag$}
\address{$^\dag$Microsoft, USA\\
$^\ddag$Fondazione Bruno Kessler and University of Trento, Italy\\
\texttt{\small spapi@fbk.eu,\{peidongwang,junkunchen,jian.xue,jinyli,yashesh.gaur\}@microsoft.com}}
\begin{document}
%
\maketitle
\copyrightnotice{979-8-3503-0689-7/23/\$31.00~\copyright2023 IEEE}
\begin{abstract}
In real-world applications, users often require both translations and transcriptions of speech to enhance their comprehension, particularly in streaming scenarios where incremental generation is necessary. This paper introduces a streaming Transformer-Transducer that jointly generates automatic speech recognition (ASR) and speech translation (ST) outputs using a single decoder. To produce ASR and ST content effectively with minimal latency, we propose a joint token-level serialized output training method that interleaves source and target words by leveraging an off-the-shelf textual aligner. Experiments in monolingual (it-en) and multilingual (\{de,es,it\}-en) settings demonstrate that our approach achieves the best quality-latency balance. With an average ASR latency of 1s and ST latency of 1.3s, our model shows no degradation or even improves output quality compared to separate ASR and ST models, yielding an average improvement of 1.1 WER and 0.4 BLEU in the multilingual case.
\end{abstract}
\section{Introduction}
\label{sec:intro}
In many real-world applications such as lectures and dialogues, automatic speech recognition (ASR) and translation (ST) are often both required to help the user understanding the spoken content \cite{hsiao06_interspeech}. For instance, a person can have partial knowledge of the uttered language and a good knowledge of the translation language, therefore consulting the translation only when the transcription is not fully comprehended~\cite{Fuegen2009_1000013594}. 
Moreover, the consistency between transcriptions and translations represents a desirable property for speech applications~\cite{sperber-etal-2020-consistent,karakanta-etal-2021-flexibility}, and having access to both source and target texts is also particularly useful for explainable AI \cite{stahlberg-etal-2018-operation}.

Despite these requests and the several research efforts towards developing systems that are able to produce both outputs~\cite{Dong2020ConsecutiveDF,le-etal-2020-dual,xu-etal-2022-joint}, little research has focused on the streaming scenario \cite{chen-etal-2021-direct} where these outputs have to be generated while incrementally receiving additional speech content. In particular, only Weller et al., 2021 \cite{weller-etal-2021-streaming} proposed a unified-decoder solution for real-time applications that, however, leverages a fully attention-based encoder-decoder (AED) architecture~\cite{10.5555/3295222.3295349}, which is theoretically not well suited for the streaming scenario \cite{li2022recent}, and adopts the re-translation approach \cite{niehues18_interspeech}, which is well-known to be affected by the flickering problem \cite{arivazhagan-etal-2020-translation}.

\begin{figure}
    \centering
    \includegraphics[width=0.48\textwidth]{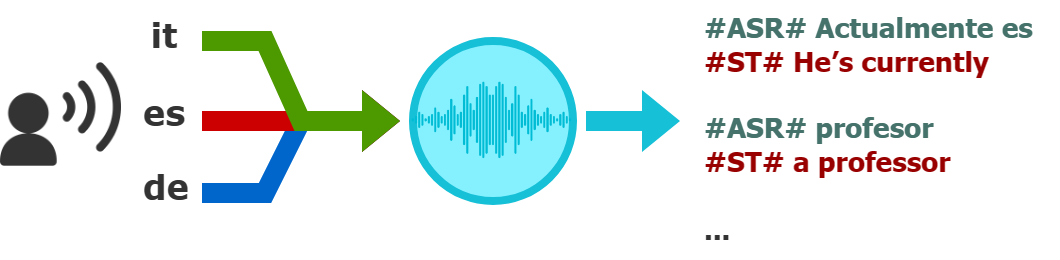}
    \caption{Illustration of the multilingual joint t-SOT with ASR and ST outputs.}
    \label{fig:jointSOT}
\end{figure}

Recently, Wang et al. 2023 \cite{wang2023lamassu} proposed a streaming language-agnostic multilingual speech recognition and translation model using neural transducers (LAMASSU), which is capable of generating both ASR and ST results. More specifically, LAMASSU with a unified prediction and joint network (LAMASSU-UNI) uses language identification (LID) information to replace the start-of-sentence token. However, in order to perform ASR and ST simultaneously, LAMASSU requires two decoder instances.

In this paper, we introduce the first streaming Transformer-Transducer (T-T) \cite{yeh2019transformer, zhang2020transformer, xiechen2021tt} able to jointly generate both transcriptions and translations using a single decoder (Figure \ref{fig:jointSOT}). To effectively learn how to produce the interleaved ASR and ST words, we propose a joint token-level serialized output training~(t-SOT) \cite{kanda22_interspeech} method that leverages an off-the-shelf neural textual aligner to build the training data without any additional costs.

Monolingual (it-en) and multilingual (\{de,es,it\}-en) experiments demonstrate the effectiveness of our proposed alignment-based joint t-SOT model, achieving the best quality-latency trade-off across languages. With an average latency of 1s for ASR and 1.3s for ST, our model not only improves the output quality compared to separate ASR and ST models, resulting in an average improvement of 1.1 WER and 0.4 BLEU  in the multilingual case, but also enables a more interpretable ST, assisted by the corresponding generated ASR outputs. Furthermore, the ability of our system to consolidate multiple tasks and languages into a single model significantly reduces the number of required systems (from 6 to 1 in the multilingual case), thus moving towards a more environmentally-friendly AI (Green AI) approach \cite{greenai}.

\begin{table*}[!t]
    \centering
    \sffamily
    \small
    \begin{tabular}{ll}
       \textbf{Transcription:}  & \asr{Ich brauche das wirklich.} \\
       \textbf{Translation:}  & \st{I really need it.} \\
       \textbf{INTER 0.0:} & \asr{\#ASR\# Ich brauche das wirklich.} \st{\#ST\# I really need it.} \\
       \textbf{INTER 1.0:} & \st{\#ST\# I really need it.} \asr{\#ASR\# Ich brauche das wirklich.} \\
       \textbf{INTER 0.5:} & \asr{\#ASR\# Ich} \st{\#ST\# I} \asr{\#ASR\# brauche} \st{\#ST\# really} \asr{\#ASR\# das} \st{\#ST\# need} \asr{\#ASR\# wirklich.} \st{\#ST\# it.}  \\
       \textbf{INTER ALIGN:} & \asr{\#ASR\# Ich} \st{\#ST\# I} \asr{\#ASR\# brauche das wirklich.} \st{\#ST\# really need it.} \\
    \end{tabular}
    \caption{\textrm{Example of a German transcription and an English translation with its corresponding interleaving INTER 0.0, 1.0, 0.5, and ALIGN.}}
    \label{tab:interleaving}
\end{table*}

\section{Related Works}
The SOT \cite{kanda20b_interspeech} method was initially introduced for non-streaming overlapped ASR and later extended to its token-level version for the streaming multi-talker scenario \cite{kanda22_interspeech} and distant conversational ASR \cite{10095367}. Recently, Omachi et al., 2023 \cite{10095896} proposed a similar approach for explainable and streaming ST by incorporating interleaved post-editing annotations into the target text but exhibiting a very high latency (more than 5 seconds).\footnote{The maximum acceptable latency limit is set between 2 and 3 seconds from most works on simultaneous interpretation \cite{fantinuoli2022defining}.}

In the streaming scenario, only Weller et al., 2021 \cite{weller-etal-2021-streaming} proposed a unified decoder for generating both ASR and ST outputs based on an AED architecture and adopting re-translation. Their framework is completely different from what we propose in this paper, since our model is Transducer-based, thus having a different architecture that naturally implements the streaming capabilities. 

The original encoder of the Transducer model \cite{graves2012sequence} was composed of LSTM layers, which were later replaced by Transformer layers due to their improved performance \cite{tian19b_interspeech,9053896}. Extensive research has been conducted on the T-T model for ASR \cite{10095553,9054260,9413562,sun21c_interspeech},
with a particular focus on the streaming scenario  \cite{9413803,9414560,9054476,huang20k_interspeech}.

Although the adoption of the T-T model has been previously proposed for the streaming ST task \cite{xue22d_interspeech}, including extensions to multilingual settings \cite{xue2022weaklysupervised,wang2023lamassu} and architectural modifications \cite{liu-etal-2021-cross,tang2023hybrid}, our paper is the first introducing a streaming single encoder-single decoder T-T model that can jointly produce ASR and ST outputs with minimal latency. Furthermore, we explore the application of the t-SOT method to jointly generate ASR and ST outputs, which has not been previously investigated in prior work.

\section{Joint t-SOT based on textual alignments}
\label{sec:method}

\subsection{Joint t-SOT}
\label{subsec:jointSOT}

In this section, we provide a detailed explanation of the joint version of the t-SOT method.
To emit both transcriptions and translations given the input speech, we serialize the ASR and ST references into a single token sequence. Specifically, we introduce two special tokens $\langle asr \rangle$ and $\langle st \rangle$ to represent the task change (the transition between ASR and ST output) and concatenate the reference transcription tokens and translation tokens by inserting $\langle asr \rangle$ and $\langle st \rangle$ between utterances (either at the sentence level or within specific words). For instance, given the transcription reference $\mathbf{r_{asr}}=[r_{asr_1}, r_{asr_2}, ..., r_{asr_m}]$ and the translation reference $\mathbf{r_{st}}=[r_{st_1}, r_{st_2}, ..., r_{st_n}]$, where $m\leq\text{len}(\mathbf{r_{asr}})$ and $n\leq\text{len}(\mathbf{r_{st}})$, the corresponding joint t-SOT reference is:
\begin{equation*}
    \mathbf{r_{\text{\tiny t-SOT}}}=[\langle asr \rangle,r_{asr_1}, r_{asr_2}, ..., r_{asr_m},\langle st \rangle,r_{st_1}, r_{st_2}, ..., r_{st_n}]
\end{equation*}
If the transcription and translation utterances are divided into chunks (composed of a single or even multiple words), the concatenation process is repeated until $m=\text{len}(\mathbf{r_{asr}})$ and $n=\text{len}(\mathbf{r_{st}})$ to obtain the final $\mathbf{r_{\text{\scriptsize t-SOT}}}$.

Note that $\langle asr \rangle$ and $\langle st \rangle$ are not considered as special tokens during training: they are added directly to the vocabulary and considered as all the other tokens in the loss computation.

\subsection{Textual alignment-based joint t-SOT}
\label{subsec:align}

In proposing the AED architecture for the ASR and ST joint decoding, Weller et al., 2021 \cite{weller-etal-2021-streaming} introduced a method for interleaving transcript and translation words, controlled by the parameter $\gamma$. In particular, the next interleaved word is a transcription word if:
\begin{equation*}
    (1.0 - \gamma) * (1+\text{count}_{asr})>\gamma * (1+\text{count}_{st})
\end{equation*}
where $\text{count}_{asr}$ and $\text{count}_{st}$ represent the count of ASR and ST words generated in the target text up to that point.
The authors explored different scenarios, including corner cases such as $\gamma=0.0$, where all the transcription words are generated first, followed by all the translation words (hereinafter, INTER 0.0), and $\gamma=1.0$, where all the translation words are followed by all the transcription words (hereinafter INTER 1.0). However, these corner cases are not actually streaming for one of the two tasks, as INTER 0.0 is not streaming for ST and INTER 1.0 is not streaming for ASR. For this reason, the authors proposed to alternate one ASR word and one ST word (hereinafter, INTER 0.5), thus realizing a streaming model for both tasks.\footnote{The authors also provided results for $\gamma=0.3$, showing consistently inferior performance compared to the other strategies. We also tried to interleave more than one word at a time when adopting INTER 0.5 but it led to significantly worse results.} The switch between the two tasks is controlled by a language token, determined from learned embeddings that are summed with the word embeddings during training and predicted at test time.

\begin{figure}[!t]
    \centering
    \includegraphics[width=0.2\textwidth]{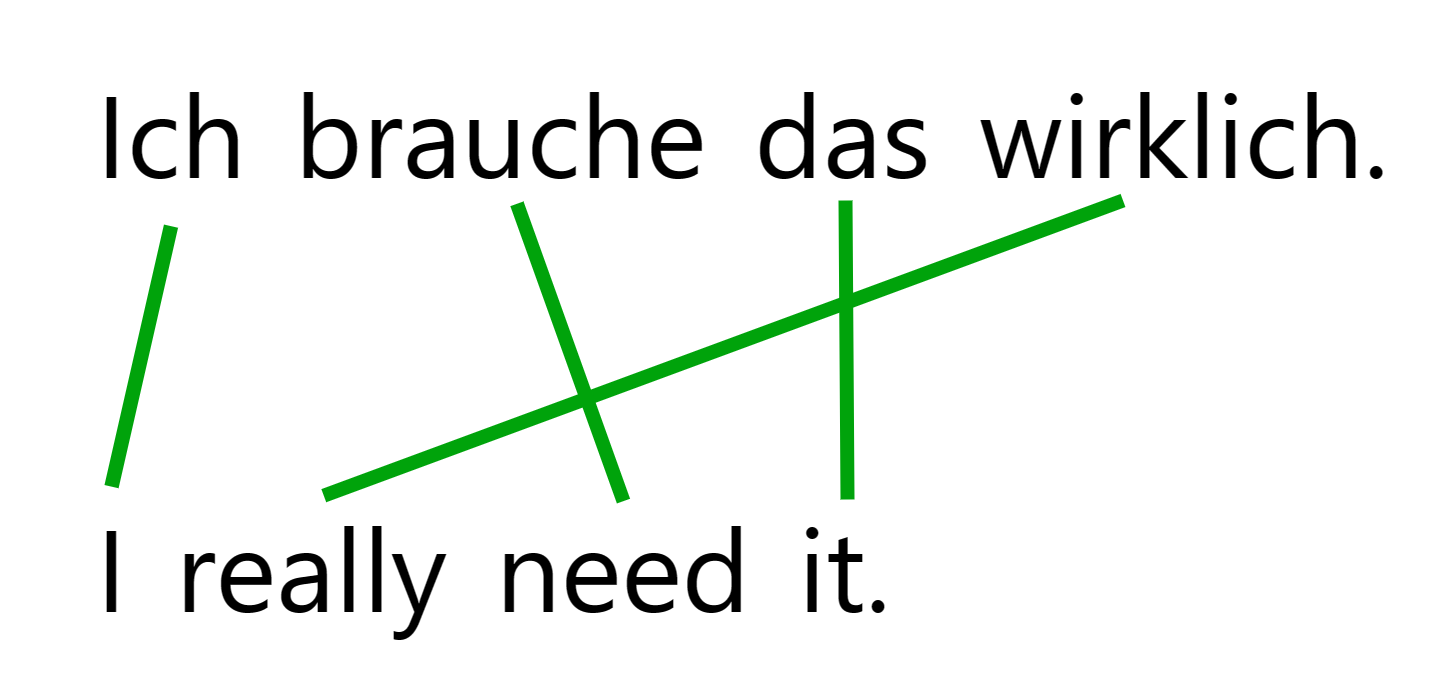}
    \caption{Example alignment between a German transcription and an English translation obtained by \texttt{awesome-align}.}
    \label{fig:align}
\end{figure}

In our approach, we first integrate the interleaving method into the t-SOT training by removing the need for learned embeddings. We replace them with specific ASR and ST tokens, as explained in Section \ref{subsec:jointSOT}, setting $\langle asr \rangle=$\enquote{\#ASR\#} and  $\langle st \rangle=$\enquote{\#ST\#}. An example of t-SOT INTER 0.0, 1.0, and 0.5 is shown in Table \ref{tab:interleaving}. 
Second, we introduce a new method for interleaving ASR and ST words based on a semantically-motivated approach. We leverage an off-the-shelf neural textual aligner \texttt{awesome-align}~\cite{dou2021word} to predict the alignment between transcription and translation texts, which are exploited to build the training data. Then, let again $\mathbf{r_{asr}}=\{r_{asr_1},...,r_{asr_m}\}$ the transcription reference and $\mathbf{r_{st}}=\{r_{st_1},...,r_{st_n}\}$ the translation reference, we build the alignment-based interleaving (hereinafter, INTER ALIGN) by applying the following rules:
\begin{enumerate}
    \item If a transcription word $r_{asr_i}$ and a translation word $r_{st_j}$ are uniquely aligned (as the words \enquote{Ich} and \enquote{I} in Figure \ref{fig:align}), they are interleaved following INTER 0.5:
    \begin{itemize}
        \item[$\Rightarrow$] $\mathbf{r_{\text{\scriptsize t-SOT}}}$ += \asr{\#ASR\#}, \asr{$r_{asr_i}$}, \st{\#ST\#}, \st{$r_{st_j}$}
    \end{itemize}
    \item If $k$ consecutive transcription words $r_{asr_i}$, $r_{asr_{i+1}}$,..., $r_{asr_{i+k-1}}$ are aligned with the same translation word $r_{st_j}$, we interleave them together as a single word (valid also in the opposite case):
    \begin{itemize}
        \item[$\Rightarrow$]  $\mathbf{r_{\text{\scriptsize t-SOT}}}$ += \asr{\#ASR\#}, \asr{$r_{asr_i},r_{asr_{i+1}},...,r_{asr_{i+k-1}}$}, \st{\#ST\#}, \st{$r_{st_j}$} 

    \end{itemize}
    \item If a transcription word $r_{asr_i}$ is aligned with a translation word $r_{st_a}$ that appears consecutively after the current translation word $r_{st_j}$, but $r_{asr_i}$ is not also aligned with $r_{st_j}$ (as in Figure 2, where $r_{asr_i}=$\enquote{brauche} is aligned with $r_{st_a}=$\enquote{need}, but not with $r_{st_j}=$\enquote{really}), we consider all the words $r_{st_j},...,r_{st_a}$ for the interleaving (the condition must also be satisfied in the reverse direction):
    \begin{itemize}
       \item[$\Rightarrow$]  $\mathbf{r_{\text{\scriptsize t-SOT}}}$ += \asr{\#ASR\#}, \asr{$r_{asr_i}$}, \st{\#ST\#}, \st{$r_{st_j},...,r_{st_a}$} 
    \end{itemize}
    \item If a transcription word $r_{asr_{\text{miss}}}$  appears consecutively after $r_{asr_i}$ and is not aligned with any translation words $r_{st_j},...,r_{st_n}$, the word is included in the subsequent interleaving sequence (valid also in the opposite case):
    \begin{itemize}
        \item[$\Rightarrow$]  $\mathbf{r_{\text{\scriptsize t-SOT}}}$ += \asr{\#ASR\#}, \asr{$r_{asr_i}$}, \st{\#ST\#}, \st{$r_{st_j}$,...}, \asr{\#ASR\#}, \asr{$r_{asr_{\text{miss}}},...$} 
    \end{itemize}
    \item If no $\mathbf{r_{asr}}$ or $\mathbf{r_{st}}$ words are left, we concatenate together all the remaining words of, respectively, $\mathbf{r_{st}}$ or $\mathbf{r_{asr}}$:
    \begin{itemize}
        \item[$\Rightarrow$]  $\mathbf{\mathbf{r_{\text{\scriptsize t-SOT}}}}$ += \st{\#ST\#}, \st{$r_{st_j},...,r_{st_n}$} \qquad\qquad or \\  $\mathbf{r_{\text{\scriptsize t-SOT}}}$ += \asr{\#ASR\#}, \asr{$r_{asr_i},...,r_{asr_m}$}
    \end{itemize}
    
\end{enumerate}

With the transcription and translation example in Table~\ref{tab:interleaving}, we obtain the alignment shown in Figure \ref{fig:align}. Its corresponding INTER ALIGN output is shown in the last row of Table~\ref{tab:interleaving}. In particular, since \enquote{Ich} (ASR) and \enquote{I} (ST) are uniquely aligned, they are interleaved in the INTER 0.5 fashion. But, since \enquote{brauche} (ASR) is aligned with \enquote{need} (ST), and \enquote{really} (ST) is aligned with \enquote{wirklich} (ASR), the entire ASR block composed of \enquote{brauche das wirklich} is inserted before the corresponding ST words \enquote{really need it}.

\begin{table*}[!t]
\setcounter{table}{2}
    \centering
    \footnotesize
    \setlength{\tabcolsep}{3.7pt}
    \begin{tabular}{l|c||cc|cc||cc|cc||cc|cc}
       \multirow{2}{*}{\textbf{Model}} &  \multirow{2}{*}{\textbf{\# param.}} & \multicolumn{4}{c||}{\textbf{it-en}} & \multicolumn{4}{c||}{\textbf{es-en}} & \multicolumn{4}{c}{\textbf{de-en}} \\
       \cline{3-14}
       & & \textbf{WER} & \textbf{LAAL} & \textbf{BLEU} & \textbf{LAAL} & \textbf{WER} & \textbf{LAAL} & \textbf{BLEU} & \textbf{LAAL} & \textbf{WER}  & \textbf{LAAL} & \textbf{BLEU} & \textbf{LAAL} \\
       \hline
        CoVoST 2 baseline \cite{wang21s_interspeech}$^\dag$ & \multirow{2}{*}{-} & \multirow{2}{*}{27.40} & \multirow{2}{*}{-} & 0.20* & - & \multirow{2}{*}{\textbf{16.00}} & \multirow{2}{*}{-} & 12.0 & - & \multirow{2}{*}{21.40} & \multirow{2}{*}{-} & 8.4 & - \\
        \quad + English ASR pre-train$^\dag$ & & & & 11.30 & - & & & 23.00 & - & & & 17.4 & - \\
        \hline
        \multirow{3}{*}{separate ASR \& ST} & 3$\cdot$(185.1M & \multirow{3}{*}{25.83} & \multirow{3}{*}{1191} & \multirow{3}{*}{16.41} & \multirow{3}{*}{1844} & \multirow{3}{*}{22.69} & \multirow{3}{*}{1149} & \multirow{3}{*}{19.24} & \multirow{3}{*}{1682} & \multirow{3}{*}{23.11} & \multirow{3}{*}{1071} & \multirow{3}{*}{19.11} & \multirow{3}{*}{1613} \\
        & + & & & & & & & & & & & \\
        & 185.1M) & & & & & & & & & & & \\
        \hline
        \multirow{3}{*}{multilingual ASR \& ST} & 188.5M & \multirow{3}{*}{23.48} & \multirow{3}{*}{1181} & \multirow{3}{*}{21.06} & \multirow{3}{*}{1663} & \multirow{3}{*}{22.84} & \multirow{3}{*}{1147} & \multirow{3}{*}{22.76} & \multirow{3}{*}{1622} & \multirow{3}{*}{21.82} & \multirow{3}{*}{1133} & \multirow{3}{*}{21.51} & \multirow{3}{*}{1642} \\
         & + & & & & & & & & & & & \\
         & 185.1M & & & & & & & & & & & \\
        \hline
        joint t-SOT INTER 0.0 & 188.5M & 21.81 & 1228 & 20.42 & 3894 & 20.76 & 1196 & 23.26 & 3752 & 20.82 & 1168 & 21.53 & 3647 \\
        joint t-SOT INTER 1.0 & 188.5M & 26.05 & 3389 & \textbf{22.17} & 1743 & 23.45 & 3172 & \textbf{23.99} & 1683 & 26.88 & 3234 & \textbf{21.85} & 1964 \\
        joint t-SOT INTER 0.5 & 188.5M & 22.35 & 1110 & 20.22 & 1515 & 21.19 & 1126 & 22.25 & 1468 & \underline{\textbf{21.35}} & \underline{1051} & 20.19 & 1547 \\
        \hline
        joint t-SOT INTER ALIGN & 188.5M & \underline{\textbf{21.74}} & \underline{\textbf{1092}} & \underline{21.80} & \underline{\textbf{1355}} & \underline{21.04} & \underline{\textbf{1094}} & \underline{23.42} & \underline{\textbf{1341}} & 22.07 & \textbf{1043} & \underline{21.36} & \underline{\textbf{1335}} \\
    \end{tabular}
    \caption{WER$\downarrow$ and BLEU$\uparrow$ results for the multilingual setting (\{it, es, de\}-en) with their latency LAAL$\downarrow$. *Results obtained using CoVoST 2 data only for training. $^\dag$Non-streaming model. \textbf{Bold} represents overall best result, \underline{underline} represents best result balancing both quality and latency.}
    \label{tab:multi}
\end{table*}

\section{Experimental Settings}
\label{sec:exp}
We adopt a streaming T-T architecture \cite{xiechen2021tt} with 24 Transformer layers for the encoder, 6 LSTM layers for the predictor and 2 feed-forward layers for the joiner. The Transformer encoder has 8 attention heads, the embedding dimension is 512 and the feed-forward units are 4096. We use a chunk size of 1 second with 18 left chunks. The LSTM predictor has 1024 hidden units as well as the feed-forward layers of the joiner. Dropout is set to 0.1.  We use 80-dimensional log-mel filterbanks
as features, which are sampled every 10 milliseconds. Before feeding them to the Transformer encoders, we process the features with 2 layers of CNN with stride 2 and kernel size of (3, 3), with an overall input compression of 4. 

Our experiments are performed using 1k hours of proprietary data for each language (German, Italian, Spanish to English) and the models are tested on the CoVoST2 dataset \cite{wang21s_interspeech}. AdamW \cite{loshchilov2018decoupled} is used as optimizer with the RNN-T loss \cite{graves2012sequence}. The training steps are 6.4M for the joint t-SOT models and 3.2M for the separate ASR and ST models.\footnote{We noticed that a longer training of 6.4M steps does not improve or even degrades the performance.} Checkpoints are saved every 320k steps. The learning rate is set to 3e-4 with Noam scheduler, 800k warm-up steps and linear decay.
The vocabulary is based on SentencePiece \cite{kudo-richardson-2018-sentencepiece} and has dimension 4k for all the monolingual models, all the separate ASR and ST models, and the multilingual source ST model (since the target is always English). For the multilingual (source) joint t-SOT and ASR models, the vocabulary size is set to 8k. Coverage is always set to 1.0.

We use 16 NVIDIA V100 GPUs with 32GB of RAM for all the training and a batch size of 350k. We select the last checkpoint for inference, which is then converted to open neural network exchange (ONNX) format and compressed. The beam size of the beam search is set to 7. 

We report WER for the ASR output quality and BLEU\footnote{sacreBLEU \cite{post-2018-call} version 2.3.1} for the ST output quality. Latency is measured in milliseconds~(ms) with the length-adaptive average lagging (LAAL)~\cite{papi-etal-2022-generation}, which is derived from the speech adaptation~\cite{ma-etal-2020-simulmt} of the average lagging (AL) metric \cite{ma-etal-2019-stacl}, incorporating the capability to handle predictions longer than the reference.

\section{Results}
\label{sec:res}

\subsection{Monolingual Results}
Table \ref{tab:mono} presents the results of the Italian monolingual ASR, ST and joint t-SOT models. 

First, we observe how effective is the joint t-SOT compared to training separate ASR and ST models. With the only exception of the ASR task for INTER 1.0 and the ST task for INTER 0.0, the joint t-SOT models always outperform the separate ASR and ST architectures with improvements ranging from 0.63 to 1.18 WER while maintaining the same latency for ASR, and from 0.64 to 2.79 BLEU with also an average latency reduction of 312ms for ST. Therefore, the obtained results indicate the joint t-SOT as a very promising approach. 
Moreover, the high latency shown by INTER 1.0 for ASR (over 3.5s) and  INTER 0.0 for ST (approximately 3s) was expected since, for these two approaches, only one of the two modalities is actually streaming (as also already discussed in Section \ref{subsec:align}). 

Second, in contrast to Weller et al., 2021 \cite{weller-etal-2021-streaming}, we notice that INTER 0.5 achieves the best WER result instead of INTER 1.0 while, in accordance with them, the best BLEU is obtained by INTER 0.0. The lowest latency is achieved by INTER 0.5 and INTER ALIGN for ASR, and by INTER ALIGN for ST with a very large margin (between 150 and 1600ms of latency reduction). Considering both output quality and latency, the overall best result (underlined in Table \ref{tab:mono}) is obtained by INTER 0.5 for ASR, closely followed by INTER ALIGN, and INTER ALIGN for ST. Therefore, in the monolingual setting, INTER ALIGN emerges as the optimal model for jointly performing the ASR and ST tasks.

\begin{table}[!t]
\setcounter{table}{1}
    \centering
    \footnotesize
    \setlength{\tabcolsep}{1.8pt}
    \begin{tabular}{l|c||cc|cc}
       \textbf{Model}  &  \textbf{\# param.} & \textbf{WER} & \textbf{LAAL} & \textbf{BLEU} & \textbf{LAAL} \\
       \hline
        CoVoST 2 baseline \cite{wang21s_interspeech}$^\dag$ & \multirow{2}{*}{-} & \multirow{2}{*}{27.40} & \multirow{2}{*}{-} & 0.20* & - \\
        \quad + English ASR pre-train$^\dag$ & & & & 11.30 & - \\
        \hline
        \multirow{3}{*}{separate ASR \& ST} & 185.1M & \multirow{3}{*}{25.83} & \multirow{3}{*}{1191} & \multirow{3}{*}{16.41} & \multirow{3}{*}{1844} \\
        & + & & & \\
        & 185.1M & & & \\
        \hline
        joint t-SOT INTER 0.0 & 185.1M & 24.81 & 1232 & 17.05 & 2972 \\
        joint t-SOT INTER 1.0 & 185.1M & 29.69 & 3683 & \textbf{19.20} & 1734 \\
        joint t-SOT INTER 0.5 & 185.1M & \underline{\textbf{24.65}} & \underline{\textbf{1126}} & 17.58 & 1508 \\
        \hline
        joint t-SOT INTER ALIGN & 185.1M & 25.20 & 1128 & \underline{18.65} & \underline{\textbf{1355}} \\
    \end{tabular}
    \caption{WER$\downarrow$ and BLEU$\uparrow$ results for the monolingual setting (it-en) with their latency LAAL$\downarrow$. *Results obtained using CoVoST 2 data only for training. $^\dag$Non-streaming model. \textbf{Bold} represents overall best result, \underline{underline} represents best result balancing both quality and latency.}
    \label{tab:mono}
\end{table}

\begin{table*}[!ht]
\setcounter{table}{3}
    \centering
    \sffamily
    \small
    \begin{tabular}{llcl}
       \multirow{4}{*}{\#1} & \multirow{2}{*}{\textbf{Reference:}}  & \textbf{IT} & \asr{Per questo venne \textbf{martirizzato}.} \\
       &  & \textbf{EN} & \st{For that reason he was \textbf{martyred}.} \\
       & \multirow{2}{*}{\textbf{Hypothesis:}}  & \textbf{IT} & \asr{Per questo venne \underline{\textbf{utilizzato}}.} \\
       &  & \textbf{EN} & \st{That's why he was \underline{\textbf{used}}.} \\
        \hline
       \multirow{4}{*}{\#2} & \multirow{2}{*}{\textbf{Reference:}}  & \textbf{IT} & \asr{In gara unica, da disputare tra le vincenti delle \textbf{semifinali}.} \\
       &  & \textbf{EN} & \st{A single match played by those who won the \textbf{semifinals}.} \\
       & \multirow{2}{*}{\textbf{Hypothesis:}}  & \textbf{IT} & \asr{In gara unica da disputare tra i vincenti delle \underline{\textbf{finali}}}. \\
       &  & \textbf{EN} & \st{In a single match to be played among the winners of the \underline{\textbf{finals}}.} \\
       \hline
        \multirow{4}{*}{\#3} & \multirow{2}{*}{\textbf{Reference:}}  & \textbf{IT} & \asr{Più veloce persino della media degli Space \textbf{Marine}.} \\
       &  & \textbf{EN} & \st{Even faster than the average Space \textbf{Marine}.} \\
       & \multirow{2}{*}{\textbf{Hypothesis:}}  & \textbf{IT} & \asr{Più veloce persino della media degli Space \textbf{\underline{Marianne}}.} \\
       &  & \textbf{EN} & \st{Even faster than the average Space \underline{\textbf{Marianne}}.} \\
       \hline
        \multirow{4}{*}{\#4} & \multirow{2}{*}{\textbf{Reference:}}  & \textbf{IT} & \asr{Viene misurata in \textbf{Joule} nel sistema internazionale.} \\
       &  & \textbf{EN} & \st{Its measuring unit is \textbf{Joule} in the international system.} \\
       & \multirow{2}{*}{\textbf{Hypothesis:}}  & \textbf{IT} & \asr{Viene misurata in \underline{\textbf{già}} nel sistema internazionale.} \\
       &  & \textbf{EN} & \st{It is measured in \underline{\textbf{down}} in the international system.} \\
       \hline
        \multirow{4}{*}{\#5} & \multirow{2}{*}{\textbf{Reference:}}  & \textbf{IT} & \asr{In casa Porru, nella \textbf{camera dei forestieri}, c'era una donna che piangeva.} \\
       &  & \textbf{EN} & \st{In the house of the Porru family, in the \textbf{guest room}, there was a woman crying.} \\
       & \multirow{2}{*}{\textbf{Hypothesis:}}  & \textbf{IT} & \asr{In casa porru nella \underline{\textbf{stanza dei forestieri}} c'era una donna che piangeva.} \\
       &  & \textbf{EN} & \st{In the house porru in the \underline{\textbf{chamber of the strangers}} there was a woman crying.} \\
    \end{tabular}
    \caption{\textrm{Mistranslations examples with their corresponding generated transcriptions of the joint t-SOT INTER ALIGN model extracted from the Italian-English CoVoST 2 test set.}}
    \label{tab:interpretability}
\end{table*}

\subsection{Multilingual Results}
We extend our analysis to the multilingual setting by incorporating two additional source languages: Spanish, an Italic/Romance language with subject-verb-object (SVO) ordering similar to Italian, and German, a Germanic language with subject-object-verb (SOV) ordering \cite{languages}. In Table \ref{tab:multi}, we compare the joint t-SOT methods with both monolingual and multilingual ASR and ST models. 

Looking at the results of the separate ASR and ST models, we observe a significant improvement going from monolingual to multilingual, particularly for Italian and German ASR (with an improvement of, respectively, 1.29 and 2.35 WER) and for all languages in ST (with an average BLEU improvement of 3.52). 
Consistent with the findings from the monolingual experiments, our joint t-SOT methods outperform the monolingual and multilingual separate ASR and ST models considering both the output quality and the latency. 

While INTER 1.0 achieves the highest BLEU scores across all languages, it also exhibits the highest, hence worst, WER. In contrast, no clear trend emerges for the best WER results. Regarding latency, the INTER ALIGN method consistently achieves the lowest, hence best, LAAL, with an average of 1s for ASR and 1.3s for ST. Balancing both quality and latency, the overall best results are obtained by the INTER ALIGN method, with the only exception of German ASR where the WER of the INTER 0.5 is slightly better. 

In conclusion, the joint t-SOT method, and in particular the INTER ALIGN approach, proves to be the most effective solution for jointly generating ASR and ST outputs, delivering high-quality results with minimal latency. The results show that the joint t-SOT INTER ALIGN achieves significant improvements compared to the separate multilingual ASR and ST models, with an average reduction of 1.1 WER and 0.4 BLEU across all languages, while maintaining comparable or even slightly lower latency (approximately 200ms average reduction).
These findings highlight the efficiency of our proposed approach, which consolidates both ASR and ST functionalities into a single model.

\subsection{Interpretable ASR and ST Results}
To examine the relationship between the ASR and ST outputs obtained by our joint t-SOT models, we conducted a manual analysis of the generated texts. We focused on the Italian to English language pair and selected the joint t-SOT INTER ALIGN model as it resulted in the best one for the streaming scenario. Representative examples extracted from the CoVoST 2 test set are shown in Table \ref{tab:interpretability}. 

The first example shows how a wrong transcription of the verb \enquote{martirizzare} (en: \enquote{\textit{martyr}}) to the verb \enquote{utilizzato} (en: \enquote{\textit{use/utilized}}) leads to a wrong translation having the same meaning of the wrong transcription. Additionally, example 2 proves how an omission in the transcription also leads to the same omission in the translation (it: \enquote{\textit{finali}}/en: \enquote{\textit{finals}} instead of it: \enquote{\textit{semifinali}}/en: \enquote{\textit{semifinals}}). 

Examples 3 and 4 present another interesting phenomenon related to the wrong recognition of named entities and terminology. It has been previously demonstrated that failures in named entities recognition often produce the insertion of a completely different name or a common noun instead of the correct named entity \cite{gaido-etal-2022-talking}. In fact, Example 3 shows how the name \enquote{Marine} is incorrectly recognized as \enquote{Marianne} and this affects both the transcription and the translation. In Example 4, instead, the \enquote{Joule} term is misrecognized but as the common word \enquote{già}, presumably because these two words have assonance in Italian. As a consequence, the ST output is affected by the prediction of a wrong ASR word but, differently from Example 1, the translation does not reflect the meaning of the wrong word \enquote{già} but is completely random. 

Lastly, in Example 5, we observe that \enquote{stanza dei forestieri} (en: \enquote{\textit{guest room}}) is literally translated by using out-of-context terms, where \enquote{chamber} is generated instead of \enquote{room} due to both concepts being expressed by the same Italian word \enquote{stanza}. 

Therefore, by analyzing the transcriptions and translations produced by our joint t-SOT model, we can better identify and understand the root causes of mistranslations, leading to a more interpretable output. This highlights the potential of our method to leverage the generated transcription to enable explainable ST.

\section{Conclusions}
This paper introduced the first streaming Transformer Transducer that is able to jointly generate both automatic speech recognition and translation outputs using a single decoder. To effectively produce transcription and translation tokens without increased latency, we proposed a joint token-level serialized output training that leverages an off-the-shelf neural text aligner to generate the data without any additional costs. Monolingual (it-en) and multilingual (\{de,es,it\}-en) experiments proved that our proposed approach not only better balances the quality and the latency constraints of the streaming scenario, with an average latency of 1s for ASR and 1.3s for ST but also outperforms separate ASR and ST models by an average of 1.1 WER and 0.4 BLEU in the multilingual case. Moreover, it promotes a more explainable ST by exploiting the ASR outputs to better understand the root cause of the mistranslations and Green AI by significantly reducing the number of required systems.

\bibliographystyle{IEEEbib}
\bibliography{strings,refs}

\end{document}